# Multi-Mobile Robot Localization and Navigation based on Visible Light Positioning

*Yanyi Chen[1], Zhiqing Zhong[2], Shangsheng Wen[3], Weipeng Guan[3]*
[1]School of Automation Science and Engineering, South China University of Technology, Guangzhou, Guangdong, 510640, China;
[2]School of Physics and Optoelectronics, South China University of Technology, Guangzhou 510640, China
[3]School of Materials Science and Engineering, South China University of Technology, Guangzhou, Guangdong 510640, China;
\* *Correspondence:* shshwen@scut.edu.cn

**Abstract:** We demonstrated multi-mobile robot navigation based on Visible Light Positioning(VLP) localization. From our experiment, the VLP can accurately locate robots' positions in navigation.

## 1. Overview

With the rapid development of robot technology, robotic control is promising in research and commerce. Mobile robots are also widely used in a variety of environments. Especially in the automated warehouse, robots have the advantage of being cheaper and more efficient than human beings for simple and repetitive missions. However, Robotic Positioning is a fundamental ability of multi-mobile robot system. Outdoors, global positioning system(GPS) sometimes give us attenuated location information signal impacted by reflections and penetration[1]. It is not good for positioning in indoor circumstances. Indoors, compared with other interior positioning, such as Wireless Local Area Network(WLAN), Radio Frequency Identification(RFID), Ultra Wide Band(UWB), etc, Visible Light Positioning(VLP) theoretically has high resistance to electromagnetic interference because the information is transmitted to the sensor on optical signal[2].

Nowadays, VLP positioning technology has been widely used in robotic system. Since VLP localization method has been applied on robot operate system(ROS), the robustness and accuracy of robot pose positioning in the continuous change process have also been improved. VLP methods can be divided into photodetector-based (PD-based) and image sensor-based, and the latter method is more stable one[3]. In this paper, we will also use it. Nonetheless, the situation of LEDs effect the stability of mobile robot system. To optimal performance of this system based on VLP, the loosely-coupled multi sensors fusion method based on VLP eliminate the effects of LED shortage/outage situations[2]. Furthermore, VLP-based positioning methods also promoted the development of multi-robot collaboration system(MRCS). The system supported in[1](VO-MRCS) is also feasible and highly accurate. Hence it is workable that VLP can be embedded in multi-mobile robot system.

As the navigation technology in robotic system being more and more popular in recent years, the need for quality of navigation technology in robotic system has been increasing, which leads to higher accuracy requirements of positioning. Considering the advantages of VLP positioning mentioned above, we applied image sensor-based VLP positioning on multi-robot navigation system to determine the position and posture of each robots through its high-precision positioning arithmetic. And we present the high stability multi-robot localization and navigation based on VLP framework.

## 2. Innovation

We proposed a VLP navigation framework demonstration. The stability and real-time performance of each mobile robot using VLP localization method based on images in navigation were observed by host, and the framework of our experiments were verified in various environments. The main contributions of our work are as follow:
1.Based on the positioning of VLP, we designed a multi-robot positioning and navigation framework, which can accurately determine the position information and path of each robot in navigation.
2.A two-robot system is proposed as an example specifically, and the adaptability of system was adjusted and the localization algorithm was optimized correspondingly.
3.Based on the algorithms above, we demonstrated a high real-time and stability multi-robot navigation demonstration.

## 3. Description of Demonstration

In the implementation of this demonstration, we built an $832 \times 480 cm^2$ multi-robot navigation experimental platform. The multi-robot navigation system is consists of two turtlebot3 robots with built-in ubuntu16.04 Kinetic operating systems and a 220cm height intelligent LED light(Fig.1) whose diameter is 18cm. The structure of Turtlebot3 robot is shown as Fig.2. The multi sensors fusion method[2] which we used also improve the stability of system. The LiDAR is used to show the location by perceiving the boundary distance of the environment. The RES camera sensor above turtlebot3 is essential in this experiment, which is used to receive the optical information of LED to determine its position information. The LED emits optical signals containing ID of LED in the region-of-interest(ROI)[2].The ROI is set to determine a boundary of emission, helping promote images processing, which we will introduce detail later. The host of the two robots is a laptop with an ubuntu18.04 Melodic operating system, connecting via WiFi. Through RVIZ(3D visualization tool for ROS) the host can not only receive two robots' real-time location information while navigating but also set the goal of navigation by the 2D Nav Goal button. Even if the operate-system version of host is different from Turtlebot3 robots, their message is still compatible in this experiment. The whole environment of navigation was constructed by VLP-Constrained Gmapping[2] before multi-robot navigation.

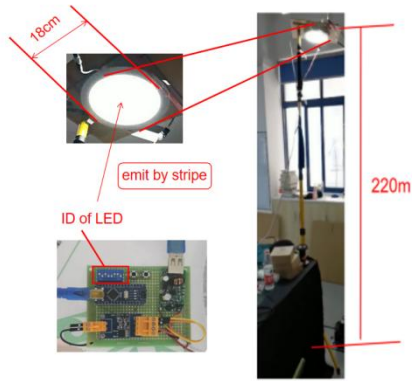

Fig.1.The structure of intelligent LED

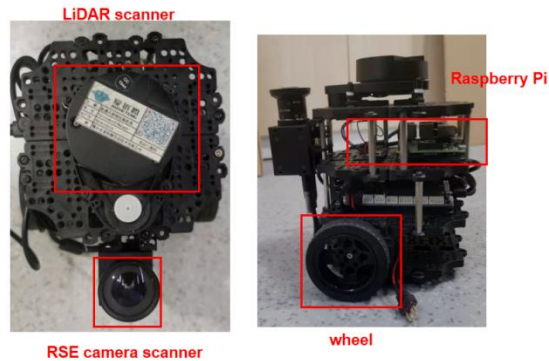

Fig.2.The structure of Turtlebot3 robot

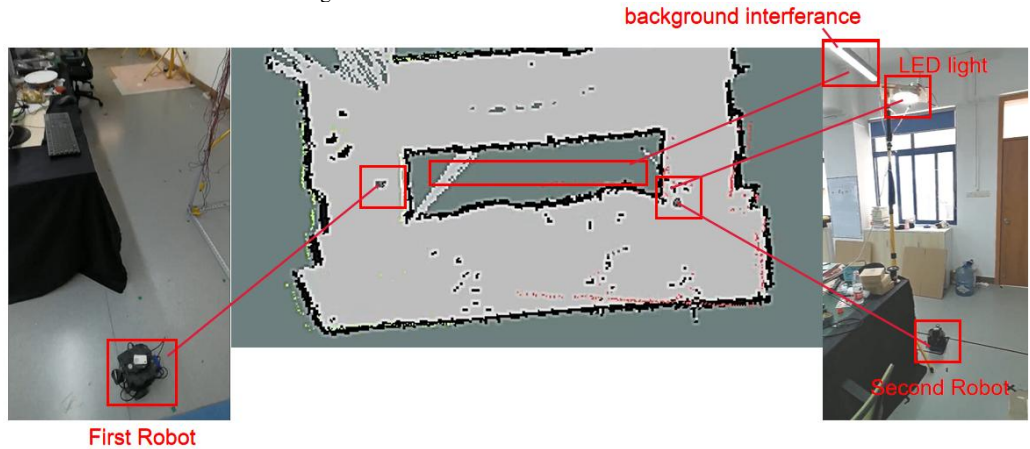

Fig.3.The experiment circumstance of dual robots Navigation and Localization based on VLP
Demo link: https://www.bilibili.com/video/BV1Df4y1T7u3?spm_id_from=333.999.0.0

For VLP, the optical signal is the medium of communication. After optical signals are emitted in the ROI of LED, they are firstly captured by the RES camera as grayscale images. Secondly the image is converted into a stripe pattern with the binarization of grayscale images. Then the stripes are decoded to the ID of LED in the host. The pose information of the robot will be calculated based on the ID of LED and the preset position information of mapping LED. Considering ambient light interference, the RSE camera also can distinguish ambient light interference by capturing the ROI. Thus it can be adapted to different environments as well.

To test the feasibility of multi-robot system based on VLP positioning, we navigated two robots with different initial positions. One was located out of the LED coverage area and the other was located in the LED coverage area. We observed the performance of VLP navigation localization while navigating the robots both staying the LED coverage area. To eliminate the error of mapping, instead measuring the distance between map boundary and LiDAR boundary, we measure the LiDAR boundary distance between two robots. In the experiment, we found that the position of the first robot is verification and correction, when it run into the LED coverage area. Besides, when two robots were both in the LED coverage area, the peak distance between the perceived boundary of the first robot and that of the second robot was less than 3.4cm, when they both found the LED. After the second robot left the LED coverage area, running straight towards the navigating destination, the peak distance between the perceived boundary and the map boundary increased to a certain extent. The delay of VLP positioning does not affect the efficiency of navigation, for the average speed of the robots into the LED coverage area is nearly the same as leaving the LED coverage area. Therefore, our experiment shows that the multi-mobile robot system framework based VLP has a greatly stable, real-time, and accurate positioning ability, after we get a correct navigation position or path.

**Acknowledgement**
This research was funded by National Undergraduate Innovation and Entrepreneurship Training Program (202110561007).